\def\preprintmode{1}
\documentclass[10pt]{article}
\ifdefined\preprintmode
\usepackage[preprint]{tmlr}
\else
\usepackage{tmlr}
\fi

\usepackage{microtype}
\usepackage{graphicx}
\usepackage{booktabs}
\usepackage{array}
\newcolumntype{L}[1]{>{\raggedright\arraybackslash}p{#1}}
\usepackage{xcolor}
\usepackage{hyperref}
\usepackage{url}

\usepackage{amsmath}
\usepackage{amssymb}
\usepackage{mathtools}
\usepackage{amsthm}
\usepackage{float}
\usepackage{algorithm}
\usepackage{algorithmic}

\usepackage{tikz}
\usetikzlibrary{positioning,arrows.meta,fit,calc}
\usepackage{pgfplots}
\pgfplotsset{compat=1.16}
\usepackage[capitalize,noabbrev]{cleveref}

\definecolor{cfblue}{HTML}{0072B2}
\definecolor{cforange}{HTML}{E69F00}
\definecolor{cfgreen}{HTML}{009E73}
\definecolor{cfred}{HTML}{D55E00}
\theoremstyle{plain}
\newtheorem{proposition}{Proposition}

\newcommand{\E}{\mathbb{E}}
\newcommand{\TV}{\mathrm{TV}}
\newcommand{\KL}{\mathrm{KL}}
\newcommand{\cF}{\mathsf{F}}
\newcommand{\cU}{\mathsf{U}}
\newcommand{\ind}{\mathbf{1}}

\title{Why Calibrated BM25 Filtering Barely Reduced Context on LongMemEval}

\ifdefined\preprintmode
\author{\name John Knight \email jknight137@tamu.edu \\
      \addr Department of Electrical and Computer Engineering\\
      Texas A\&M University, College Station, TX 77843-3128, USA
      \AND
      \name Erchin Serpedin \email eserpedin@tamu.edu \\
      \addr Department of Electrical and Computer Engineering\\
      Texas A\&M University, College Station, TX 77843-3128, USA
      \AND
      \name Hasan Kurban\thanks{Corresponding author.} \email hkurban@hbku.edu.qa \\
      \addr College of Science and Engineering\\
      Hamad Bin Khalifa University, Doha, Qatar}
\newcommand{\codelink}{\url{https://github.com/KurbanIntelligenceLab/Certfold}}

\else
\author{Anonymous authors}
\newcommand{\codelink}{\url{https://anonymous.4open.science/r/Certfold-54EE/}}
\fi

\def\month{MM}
\def\year{YYYY}
\def\openreview{\url{https://openreview.net/forum?id=XXXX}}

\begin{document}
\hypersetup{hidelinks,pdftitle={Why Calibrated BM25 Filtering Barely Reduced Context on LongMemEval}}

\maketitle
\let\AND\relax
\ifdefined\preprintmode\else
\begingroup\renewcommand{\thefootnote}{}\footnotetext{We used large language models to help write code and prepare the manuscript, to assist with analysis, and to re-grade saved model answers. \Cref{sec:human} describes how two human graders independently checked a sample of those answers. Appendix~\ref{app:impl} gives details about the tools and procedures.}\endgroup
\fi

\begin{abstract}
An assistant that summarizes old conversations may later need a detail from the original exchange. It can score archived passages against the question, but only passages that fit in the prompt reach the answering model. We tested whether a BM25 filter saves input tokens once that limit is taken into account. We calibrated the filter on passages that individually corrected an answer the model got wrong from the summaries alone. Across 75 LongMemEval questions, filtering changed one prompt and saved 0.096\% of input tokens. With and without the filter, the model answered the same questions correctly under a second automated grading. On a random sample, two human graders' settled judgments agreed with the automated grades in 92.9\% of definite comparisons (Cohen's $\kappa=0.805$). Building on earlier work that distinguishes retrieved evidence from delivered evidence, we give an exact test for packers that try whole passages in a fixed order and skip any that do not fit. Run the packer without the filter. The prompt changes exactly when an admitted passage scores below the cutoff. Code and data: \codelink.
\end{abstract}

\section{Introduction}\label{sec:intro}
An agent can summarize an old conversation to make room for new messages, then retrieve an original passage if a question calls for a missing detail \citep{sun2025folding}. Which passages should it put back? Restoring too little can leave the question unanswered. Restoring too much costs input tokens, and models can struggle to use long contexts even when the relevant passage is present \citep{liu2024lost,du2025contextlength}.

Consider a chat assistant that summarizes a user's first open-mic performance. Weeks later, the user asks how long they had watched comedy specials before performing. The summary mentions the performance but omits the interval. The original exchange contains the answer.

We call each archived passage a \emph{fold}. The answering model, or \emph{reader}, always sees the summaries. To \emph{restore} a fold, we add the entire passage to its prompt. A fold is \emph{beneficial} if restoring it alone corrects an answer the reader got wrong from the summaries.

One way to decide what to restore is to score each passage against the question. Prior work calibrates a score cutoff to retain relevant snippets \citep{chakraborty2026context}, or trains a router on questions where raw text corrected a summaries-only answer \citep{zhu2026tiermem}. A cutoff determines which passages are eligible. A second step, \emph{packing}, determines which eligible passages fit. \citet{bala2026recall} found that retrieval and delivery can differ under a token budget. We ask how much a calibrated cutoff contributes when the window already excludes some of the passages it would reject.

Our system, CertFold, ranks folds by BM25, a lexical retrieval score. We compare a \emph{recall gate}, which applies a cutoff calibrated on beneficial folds, with \emph{all-that-fits}, which applies no cutoff. Both pack whole passages in the same order. Our objectives are to measure how the cutoff changes the final prompt, explain when and why it changes delivery, and measure the token and accuracy trade-off when a policy restores fewer passages.

On 75 LongMemEval development questions \citep{wu2024longmemeval}, the policies produced identical prompts for 74 questions. On the open-mic question the cutoff excluded one passage and changed the answer, although the rubric accepted both answers. The filter saved 0.096\% of input tokens. A few beneficial passages scored low, so the cutoff admitted most candidates. The window already excluded nearly every passage the cutoff would have rejected.

A change is most likely when the window can hold most of what passes the cutoff, although a smaller window can still admit a short passage below it. In a replay that counted tokens without generating answers, savings from the original cutoff reached 7.3\% once the entire archive fit.

\textbf{Contributions.} We isolate the effect of filtering on the final prompt (\cref{sec:evidence}) and give an exact test for a change in delivery that requires one packing pass (\cref{sec:selectivity,sec:when}). We measure how more selective policies trade input tokens for correct answers at the tested window (\cref{sec:policies}). We also check how the choice of retrieval score and disagreements about which folds helped affect the cutoff (\cref{sec:scores,sec:human}).

\textbf{Organization.} \Cref{sec:certfold} defines the comparison. \Cref{sec:evidence,sec:selectivity,sec:when} follow a passage from selection to delivery and examine what changes as the window grows. \Cref{sec:policies,sec:scores,sec:human} examine tighter budgets, alternative scores and the reliability of the grades, respectively. \Cref{sec:implications} gives the costs and limits, \cref{sec:related} compares prior approaches, and \cref{sec:conclusion} explains how to apply the packing test elsewhere.

\section{Archive, policies and outcomes}\label{sec:certfold}
This section describes how we isolated the effect of the cutoff by holding the archive, reader and packing rule fixed.

\textbf{Data.} Each question in the cleaned LongMemEval-S release comes with a history of dated chat sessions \citep{wu2024longmemeval}. We selected 75 questions before seeing model outputs. We excluded abstention questions and histories sharing a session with any question used before this study. We took the rest in hashed order, discarding a question if its history contained an answer session for a question already selected, or if a selected history contained one of its answer sessions. Shared background sessions still link 74 of the 75 questions. We used these same development questions to design and evaluate the comparison. We grouped whole sessions chronologically into chunks of up to 8,192 Qwen3 tokens, keeping any longer session as its own chunk. The result was 15 to 18 chunks per question and 1,192 question--chunk pairs.

\textbf{What the reader sees.} Before seeing the question, a Qwen3-8B writer (revision \texttt{b968826}) had up to 256 tokens per chunk to summarize facts, names, dates, events, preferences and corrections. Every policy kept these summaries in the prompt. A Qwen3-14B reader (revision \texttt{40c0698}) received the packed history, current date and question. It was asked to extract the relevant information and answer step by step. Decoding was greedy, with thinking disabled and a 512-token answer cap. \Cref{app:impl} gives the prompt specifications.

\textbf{What counts as beneficial.} We label a question--chunk pair positive if restoring that chunk alone corrects a summaries-only error. A chunk can contain the answer without changing the reader's response, and two chunks may help jointly. We therefore grade the final answer separately after packing.

\textbf{Four policies.} Folded-only uses the summaries alone. Top-two considers the two highest-scoring folds, the recall gate considers all folds at or above its cutoff, and all-that-fits considers every fold. Each restoration policy tries candidates in descending BM25 order, breaking ties by archive order. The packer inserts each candidate in chronological order alongside the summaries and previously admitted passages. It keeps the candidate if the whole prompt fits with the answer reserve. Otherwise it skips that candidate and tries the next. Every answer-generating run used a 32,768-token window.

\textbf{Two scores.} BM25 scores each fold against the question and its date (\cref{app:rank}). We later scored the same folds with BGE, a dense embedding model \citep{xiao2023cpack}. All main policy comparisons use BM25 packing order, including those where BGE sets the cutoff. \Cref{app:bgegate} separately checks BGE packing order.

\textbf{Calibration.} We randomly ordered the 1,192 question--chunk pairs, used the first 600 to calibrate the gate and reserved the next 200 for audit. The remaining 392 were unlabeled. Under the original grades, 37 calibration folds were positive. At the prespecified miss level $\alpha=0.1$, the rank rule gives $\lfloor0.1(37+1)\rfloor=3$. The gate selected every fold scoring at least as high as the third-lowest positive, 1.560604. We froze this cutoff before observing the audit outcomes. We call it the \emph{executed cutoff}. Under exchangeability of score/label pairs, the rule bounds the probability of excluding a positive at selection, averaged over calibration and a new draw (\cref{app:rank}). Packing may still drop a selected fold. All four policies answered the same questions, including those that contributed calibration folds. \Cref{app:rq4complete} describes the reuse of earlier answers.

\textbf{Grading.} The original reader graded its own answers against question-specific rubrics. Four explicit nonanswers had been graded correct, so we re-graded all 877 distinct requests (951 saved answer occurrences) with OpenAI's \texttt{gpt-6.1-terra} under the same rubrics. A second model adjudicated 26 selected cases. The review packets withheld the original grades, scores, policy and split. The review produced 872 definite grades and five unresolved cases. Of 94 definite grades that changed, 84 changed from incorrect to correct (\cref{app:rq4complete}). We call the resulting labels \emph{re-graded labels}. A \emph{re-graded cutoff} is one recomputed from those labels. The original prompts and answers remain available for comparison. Unless otherwise identified, ranges on label-dependent results span assignments of the unresolved labels.

Two human graders independently checked a blinded sample (\cref{sec:human}). \Cref{app:impl} gives the grading procedure. \Cref{sec:implications} discusses the missing records for the four nonanswers.

With these definitions, the gate and all-that-fits differ only in whether the cutoff is applied before packing. \Cref{sec:evidence} reports what that changed.

\section{What reached the reader}\label{sec:evidence}
We first compare the prompts and answers with and without the cutoff. We then use the audit folds to locate any beneficial passages lost during selection or packing.

The gate selected 1,096 of 1,192 candidates, but the packer delivered only 291 chunks. Removing the cutoff and considering all 1,192 candidates delivered 292. The two policies produced exactly the same prompt for 74 of 75 questions. On the open-mic question, all-that-fits delivered one extra chunk and answered ``two months,'' the reference answer. The gate answered ``three months.'' The rubric for temporal questions accepts a one-month error, so both answers count as correct. This was the only question on which the extra filtering could have changed the paired result. The chunk the gate left out is an audit fold labeled non-beneficial.

\begin{table}[ht]
\caption{Filtering removed one passage from the final prompts across 75 questions. Selected and Delivered count folds across questions. Correct counts questions under the re-graded labels. Recovery counts corrections among 52 summaries-only errors, and Damage counts new errors among 23 summaries-only successes. Input is mean prompt tokens per question.}
\label{tab:rq4utility}\centering\footnotesize\setlength{\tabcolsep}{3pt}
\begin{tabular}{lrrrrrr}\toprule
Policy & Selected & Delivered & Correct & Recovery & Damage & Input\\\midrule
Folded only & 0 & 0 & 23 & 0 & 0 & 5,397\\
Top-two & 150 & 150 & 47--48 & 27--28 & 3 & 19,090\\
Recall gate & 1,096 & 291 & 50 & 30 & 3 & 30,423\\
All-that-fits & 1,192 & 292 & 50 & 30 & 3 & 30,452\\\bottomrule
\end{tabular}\end{table}

Restoration helped relative to summaries alone (\cref{tab:rq4utility}). The gate and all-that-fits each corrected 30 of 52 baseline errors and introduced three errors on the 23 baseline successes. They agreed on correctness for every question. Across the cohort, the gate saved 2,186 of 2,283,913 input tokens (0.096\%). That is before the cost of labeling the calibration and audit folds, which was 651 restored-fold answers and their grades (\cref{app:rq4complete}). Top-two used substantially less context. The gate used 59.37\% more input than top-two for two to three additional correct answers in aggregate. These two policies delivered different amounts of evidence, so the difference reflects how much was restored as well as how it was chosen.

\begin{table}[ht]
\caption{Selection and delivery on the 200 audit folds. There are 12 definite positives and one unresolved fold. All three policies deliver that fold, so the miss counts are the same whatever its grade.}
\label{tab:rq4audit}\centering\footnotesize\setlength{\tabcolsep}{4pt}
\begin{tabular}{lrrr}\toprule
Stage or policy & Misses & Negatives kept & Total\\\midrule
Gate: selected & 1 & 177--178 & 189\\
Gate: delivered & 4 & 39--40 & 48\\
Top-two: delivered & 6 & 16--17 & 23\\
All-that-fits: delivered & 4 & 40--41 & 49\\\bottomrule
\end{tabular}\end{table}

Of the four beneficial audit folds the gate failed to deliver, one scored below the cutoff and three passed it but were skipped during packing (\cref{tab:rq4audit}). All-that-fits missed the same four. Top-two missed six while delivering fewer than half as many audit folds as the gate. Under the original grades the gate selected all nine audit positives. Its 10\% target concerns the marginal probability of missing a positive at selection (\cref{app:rank}). In the calibration set, even all-that-fits delivered only 12 of the 34 to 36 folds labeled beneficial (\cref{app:bgegate}). This distinction between selection and delivery also arises in budgeted retrieval \citep{bala2026recall}.

The calibrated cutoff therefore changed almost nothing that reached the reader, and most of the beneficial folds that went missing were lost in packing. \Cref{sec:selectivity} asks why the cutoff rejected so few candidates, and \cref{sec:when} why even those rejections rarely reached the prompt.

\section{Why the cutoff removed so little}\label{sec:selectivity}
Why did the gate reject only 96 of 1,192 candidates? A recall cutoff depends on the lowest scores among the beneficial folds. We examine those scores and how the cutoff changes with the grades and the allowed miss rate.

\textbf{Ranking quality.} The area under the receiver operating characteristic curve (AUROC) measures how often a positive outranks a negative, with ties counted as half. Under the original grades, BM25 had an AUROC of 0.56 on calibration and 0.82 on audit. The audit estimate rested on just nine positives, compared with 37 on calibration. Re-grading gave a similar pattern (\cref{tab:bgegate}). The small audit sample leaves the difference between splits uncertain.

\begin{figure}[ht]
\centering
\begin{tikzpicture}
\begin{axis}[width=0.74\textwidth,height=5.8cm,scale only axis=false,
xmin=0,xmax=100,ymin=0,ymax=38,xtick={0,25,50,75,100},ytick={3,10,20,30,36},
xlabel={Candidates scoring strictly below the positive (\%)},ylabel={Positive rank},
label style={font=\footnotesize},tick label style={font=\footnotesize},
grid=major,grid style={black!12},axis line style={black!70},
clip=false]
\addplot[only marks,mark=*,mark size=1.6pt,color=black!85] coordinates {(1.17450,1) (3.77517,2) (9.22819,3) (10.15101,4) (13.50671,5) (27.34899,7) (29.44631,9) (30.78859,10) (31.54362,11) (38.00336,12) (50.50336,13) (51.34228,14) (52.26510,15) (56.87919,16) (57.29866,17) (58.22148,18) (59.89933,19) (66.10738,20) (66.35906,21) (68.28859,22) (68.79195,23) (69.79866,24) (75.41946,25) (77.60067,26) (78.10403,27) (86.24161,28) (88.42282,29) (93.62416,30) (94.96644,31) (98.23826,32) (98.40604,33) (98.48993,34) (99.07718,35) (99.91611,36)};
\addplot[only marks,mark=o,mark size=1.9pt,color=black!85] coordinates {(27.26510,6) (28.02013,8)};
\draw[cfblue,very thick] (axis cs:0,3)--(axis cs:9.22819,3);
\addplot[only marks,mark=*,mark size=2.4pt,cfblue] coordinates {(9.22819,3)};
\node[anchor=west,font=\footnotesize,text=cfblue] at (axis cs:24,4) {Cutoff: 9.23\% below};
\draw[cfblue,thin] (axis cs:23,4)--(axis cs:9.22819,3);
\end{axis}
\end{tikzpicture}
\caption{The third-lowest positive sets the re-graded cutoff. Only 9.23\% of all candidates score strictly below it. The vertical axis ranks positive folds by BM25 score. Open circles are unresolved folds, shown as positive. Excluding either shifts higher ranks down by one and leaves the cutoff unchanged. Under the original grades, 8.05\% scored below the cutoff.}
\label{fig:tail}
\end{figure}
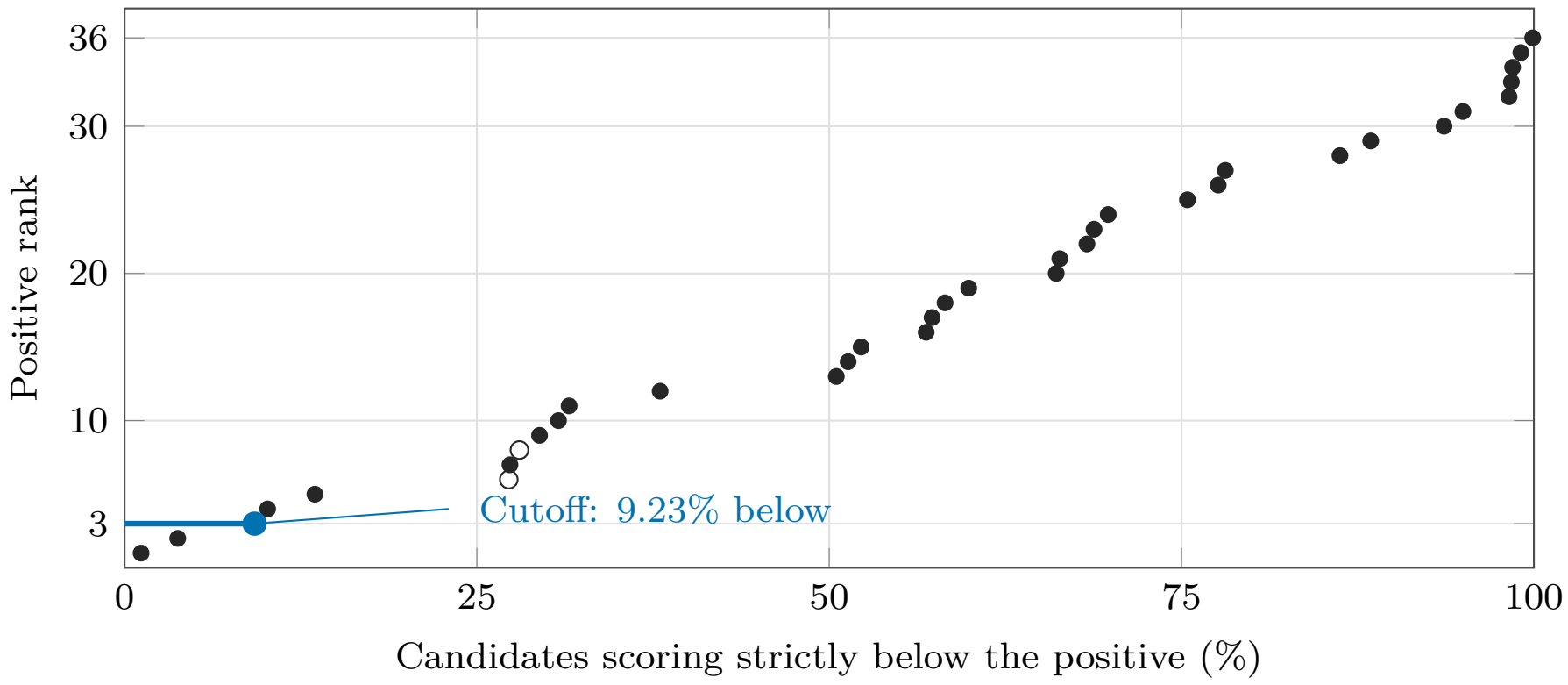

\textbf{Where the cutoff falls.} AUROC averages over all positive--negative pairs. It does not locate the third-lowest positive, which set the cutoff under both gradings.
Even after re-grading raised the cutoff, more than nine in ten candidates scored at or above it (\cref{fig:tail}).

\textbf{Sensitivity to grading.} Raising the BM25 cutoff from 1.560604 to 1.735797 left all 75 delivered prompts unchanged. All four assignments of the two unresolved calibration labels gave the same new cutoff. Substituting human grades also changed little at delivery (\cref{sec:human}). These comparisons used saved scores. The original calibration guarantee applies to the cutoff fixed before the audit.

\textbf{When would it filter more?} In the re-graded data, the first cutoff to reject at least half the pool had the same score under all four label assignments. Its rank ranged from 11 to 13, requiring a miss level of 0.31 to 0.35. Even with only 590 folds selected, 57 of 75 prompts still matched all-that-fits. More calibration positives would give finer ranks. Holding their empirical score distribution fixed, the large-sample calculation at $\alpha=0.1$ still selects 89.85\% of candidates (\cref{fig:selectivity,tab:alphasensitivity}).

Once the cutoff is set, the remaining question is which rejected passages would actually fit.

\section{When a calibrated cutoff changes the prompt}\label{sec:when}
Would the filter save more if the window were larger, or if fewer archived passages competed for space? We replayed the packing decisions using saved scores and counted the resulting input tokens. We first reproduced every prompt and both token totals from the original comparison. The other window sizes represent hypothetical capacities. Answer accuracy was measured only at the original window.

\textbf{An exact condition.} Run the packer once without the cutoff and inspect the folds it admits. The prompt changes exactly when at least one admitted fold scores below the cutoff. To see why, follow the packer in order. Removing a candidate it would skip leaves the admitted set unchanged for the next decision. If all removed candidates would have been skipped, the final set is identical. Conversely, if the cutoff removes an admitted fold, that fold is missing from the new prompt.

This argument assumes a fixed order and a packer that skips a whole fold when it fails to fit and continues to the next one. A packer that stops at the first such fold can behave differently. The test matched every replay in \cref{app:window}. A change can even lengthen the prompt: when the filter uses a different score from the packing order, excluding one fold can make room for a larger one. This happened in 60,816 of the 4,687,200 replay combinations, which include repeated archive designs.

\begin{table}[ht]
\caption{Recall gates at $\alpha=0.1$ against all-that-fits at different context windows, under the re-graded labels. Token accounting from saved scores and the packer of the executed run. Changed counts prompts of 75 that differ from all-that-fits at the same window; Delivered counts restored folds; Saved is the reduction in input tokens. Both gates pack in BM25 order. Only the 32,768-token row has answers: the BM25 gate ran as a policy, and the prompts that the BGE gate changes were answered in the additional run of \cref{sec:policies}. At that window the re-graded BM25 cutoff delivers the same prompts as the executed gate. \Cref{tab:windowfull} lists every window.}
\label{tab:window}\centering\footnotesize\setlength{\tabcolsep}{4pt}
\begin{tabular}{l rrr rrr}\toprule
 & \multicolumn{3}{c}{BM25 cutoff} & \multicolumn{3}{c}{BGE cutoff}\\
Window (tokens) & Changed & Delivered & Saved & Changed & Delivered & Saved\\\midrule
16,384 & 4 & 98 & 1.193\% & 10 & 95 & 1.237\% \\
32,768 & 1 & 291 & 0.096\% & 20 & 285 & 1.266\% \\
65,536 & 5 & 645 & 0.545\% & 37 & 621 & 3.684\% \\
98,304 & 17 & 976 & 2.415\% & 46 & 880 & 11.325\% \\
131,072 and above & 51 & 1,082 & 8.407\% & 50 & 966 & 17.662\% \\\bottomrule
\end{tabular}\end{table}

\begin{figure}[ht]
\centering
\includegraphics[width=.92\textwidth]{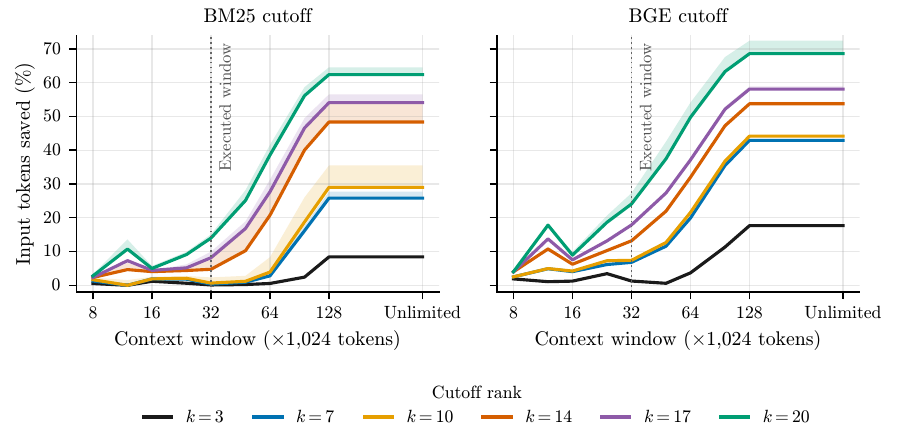}
\caption{Input tokens saved by a recall gate against all-that-fits as the context window grows, for cutoff ranks $k$; rank 3 is $\alpha=0.1$. Token accounting from saved scores and the packer of the executed run. Shading covers the unresolved calibration labels. Both panels pack in BM25 order. The dotted line marks the executed window.}
\label{fig:window}
\end{figure}

\textbf{Window size.} With the BM25 cutoff used in the original run, token savings reached 7.30\% once every question's archive fit. Recomputing the cutoff from the re-graded labels gave 8.41\% at that capacity. Up to 65,536 tokens the two cutoffs produced identical counts (\cref{tab:windowfull}). The BGE cutoff saved more when the full archive fit, but savings did not rise at every step: a larger window can admit a different combination of whole passages (\cref{tab:window,fig:window}).

\textbf{Archive size.} At the original window, restricting retrieval to two or four candidate chunks per question produced BM25 token savings between 3.5\% and 10.0\% across the tested subsets. With the full archives, each containing 15 to 18 chunks, savings fell to 0.096\%. Only the passages eligible for restoration changed. We kept every summary and the cutoff calibrated on the full archive. \Cref{tab:archivesize} gives the results for each subset size.

\textbf{A shortcut based on length.} Predict a change if the filter rejects a passage and the window has more room for restoration than the total length of the passing passages. This rule agreed with the exact test in about nine out of ten replay cases (\cref{tab:predictors}). It can miss a short rejected passage that fits after the packer skips a longer eligible one.

At the original window, almost nothing below the recall cutoff would have reached the prompt. We next tested policies that discard more.

\section{Policies that select less}\label{sec:policies}
How many tokens can a more selective policy save, and what happens to its answers? We tested two alternatives at the original window. \emph{Budget mode} calibrates on non-beneficial folds. A fold passes only if few of these calibration examples score at least as high. Under the sampling conditions of \cref{prop:conformal}, a fold known to be non-beneficial is selected with probability at most $\alpha$. Beneficial folds can also be lost. \emph{Top-$k$} simply considers the $k$ highest-scoring folds.

We chose the policy grid and the rows shown in \cref{tab:policies} after inspecting token counts and earlier grades, but before generating the new answers. The table shows nine policies alongside two saved baselines. \Cref{app:allpolicies} gives all 60 distinct selections, including settings where the chosen folds depend on an unresolved label.

The same reader answered 321 new prompts. Two prompts yielded the same answer to the same question, giving 320 distinct grading requests. Automated grading left ten unresolved. \Cref{app:impl} gives the reproduction and grading checks.

\textbf{A quarter fewer tokens.} At $\alpha=0.2$, BM25 budget mode used 25.66\% fewer input tokens than all-that-fits. It corrected five answers that all-that-fits got wrong and missed three that all-that-fits got right. The resulting counts were 52 and 50 correct out of 75, with no unresolved pairs.

\begin{table}[H]
\caption{Token savings and correct answers at the 32,768-token window. Changed counts prompts that differ from all-that-fits. Saved is the reduction in total input tokens against that baseline. Correct counts questions out of 75, with ranges for unresolved grades. The re-graded BM25 recall cutoff delivered the same prompts as the executed cutoff.}
\label{tab:policies}\centering\footnotesize\setlength{\tabcolsep}{4pt}
\begin{tabular}{lrrrrl}\toprule
Policy & Selected & Changed & Delivered & Saved & Correct\\\midrule
All-that-fits & 1,192 & 0 & 292 & 0\% & 50\\
Recall gate, BM25, $\alpha=0.1$ & 1,082 & 1 & 291 & 0.10\% & 50\\
Recall gate, BGE, $\alpha=0.1$ & 966 & 20 & 285 & 1.27\% & 49\\
Top-3 & 225 & 65 & 224 & 15.65\% & 49\\
Budget, BM25, $\alpha=0.2$ & 271 & 58 & 191 & 25.66\% & 52\\
Top-two & 150 & 75 & 150 & 37.31\% & 47--48\\
Budget, BGE, $\alpha=0.1$ & 144 & 70 & 121 & 46.66\% & 45--46\\
Budget, BM25, $\alpha=0.1$ & 136 & 68 & 113 & 48.91\% & 39--40\\
Top-1 & 75 & 75 & 75 & 59.71\% & 35--38\\
Budget, BM25, $\alpha=0.05$ & 71 & 73 & 69 & 61.88\% & 39--40\\
Folded only & 0 & 75 & 0 & 82.28\% & 23\\\bottomrule
\end{tabular}\end{table}

\textbf{Roughly half as many tokens.} At $\alpha=0.1$, BM25 answered 39 to 40 questions correctly and BGE answered 45 to 46, against 50 for all-that-fits. On definite pairs, BM25 gained four correct answers and lost 14 relative to all-that-fits. BGE gained eight and lost 12. Each comparison had one unresolved pair. For 21 questions BM25 selected no fold at all, leaving the reader only the summaries.

These are paired counts on development questions with overlapping histories. They leave accuracy on new questions open. BGE budget mode at $\alpha=0.3$ had the highest observed count in the grid, but choosing the best of 60 policies on these same questions favors chance gains (\cref{tab:allpoliciesbge}).

Tighter policies changed both the amount and the choice of evidence. We next asked how much the choice of score mattered for the recall gate.

\section{A separate test of score ranking}\label{sec:scores}
Does a better ranking help a recall gate discard more passages? We compared five scores on a panel of 150 folds, then tested BGE on the full archive.

\textbf{Which folds helped?} For each question, a fixed hash chose one fold containing a benchmark-designated answer session and one without. Half the panel therefore contained an annotated source. In the full archive, only 140 of 1,192 folds did. We measured \emph{sampled benefit} by drawing eight answers from the summaries alone and eight with the fold restored. A fold qualified if restoration raised the fraction correct by at least 0.5. Of 1,800 sampled answers, 31 grades remained unresolved after re-grading. This left 40 to 43 beneficial folds, including three to four without an annotated answer session.

\textbf{Which scores found them?} Alongside BM25 and BGE, we tested a yes-or-no model judgment and two scores based on answers sampled from the summaries alone. $Z$ adds up the changes in next-token probabilities when a fold is restored. $\Lambda$ uses the change in the probability of the entire sampled answer. In expectation, $Z$ bounds the magnitude of the change in expected reward, which can be a gain or a loss (\cref{app:commit}).

\begin{table}[ht]
\caption{Ranking and selection on the 150-fold panel. Each score uses the panel's own positives to set its recall cutoff at $\alpha=0.1$. Ranges span the unresolved benefit labels. BGE ranks sampled benefit better than BM25 here, yet selects a larger share of the panel.}
\label{tab:rq3select}
\centering\footnotesize\setlength{\tabcolsep}{3pt}
\begin{tabular}{l rrrr}
\toprule
& \multicolumn{2}{c}{Source presence} & \multicolumn{2}{c}{Sampled benefit, re-graded} \\
Score & AUROC & Selected & AUROC & Selected \\
\midrule
$Z$ & 0.8382 & 78.0\% & 0.8360--0.8556 & 48.7\% \\
BM25 & 0.9109 & 61.3\% & 0.7822--0.8042 & 61.3--68.7\% \\
BGE & 0.9340 & 59.3\% & 0.8049--0.8269 & 69.3--70.0\% \\
Need judge & 0.7751 & 100\% & 0.6875--0.7131 & 100\% \\
$\Lambda$ & 0.9376 & 58.7\% & 0.9060--0.9228 & 39.3--48.7\% \\
\bottomrule
\end{tabular}
\end{table}
BGE ranked beneficial folds somewhat better than BM25 but selected more candidates (\cref{tab:rq3select}). Its lowest-scoring positives still set a low cutoff. The probability scores $Z$ and $\Lambda$ each ranked benefit better than BM25 and selected fewer candidates. The yes-or-no judge gave its lowest score to at least one positive, so every candidate passed.

The ordering also depended on grading. BM25 ranked ahead of $Z$ under the original grades, and behind it after re-grading. The sampling interval crossed the benefit threshold for every fold labeled beneficial. \Cref{app:rq3complete} gives the sampling analysis, results within questions and four additional scores.

\textbf{Does BGE change delivery on the full archive?} We could compute BGE for every fold. The probability scores were available only on the panel. For this retrospective comparison we changed the score that set the cutoff and kept the original BM25 packing order.

\begin{table}[ht]
\caption{Full-archive recall gates at $\alpha=0.1$. Changed, Saved and Correct use the baseline and units of \cref{tab:policies}. The original BM25 gate generated answers in the main run. The re-graded BM25 gate delivered identical prompts. The additional run in \cref{sec:policies} answered every changed prompt for both BGE cutoffs. AUROC ranges span unresolved labels.}
\label{tab:bgegate}
\centering\footnotesize\setlength{\tabcolsep}{3pt}
\begin{tabular}{llrrrrrrr}
\toprule
& & \multicolumn{2}{c}{AUROC} & & & & & \\
Score & Labels & Calibration & Audit & Selected & Changed & Delivered & Saved & Correct \\
\midrule
BM25 & original & 0.5604 & 0.8197 & 1,096 (91.95\%) & 1 & 291 & 0.096\% & 50 \\
BM25 & re-graded & 0.5840--0.6026 & 0.7793--0.7988 & 1,082 (90.77\%) & 1 & 291 & 0.096\% & 50 \\
BGE & original & 0.5806 & 0.6835 & 1,125 (94.38\%) & 8 & 289 & 0.350\% & 49--50 \\
BGE & re-graded & 0.6675--0.6824 & 0.7385--0.7606 & 966 (81.04\%) & 20 & 285 & 1.27\% & 49 \\
\bottomrule
\end{tabular}
\end{table}

With the re-graded labels, BGE selected 116 fewer folds than BM25 (\cref{tab:bgegate}). Yet it delivered only seven fewer folds than all-that-fits and saved 1.27\% of input tokens. Under both gradings, BGE and all-that-fits delivered the same number of beneficial folds in the calibration and audit sets (\cref{app:bgegate}). The exact cutoff depends on a handful of low-scoring positives, which motivated a human check of their grades.

\section{Human check of the automated grades}\label{sec:human}
How closely did humans agree with the automated grader, and could a disagreement change a cutoff? Two graders independently reviewed one shuffled packet of 125 requests. It combined a simple random sample of 100 of the 877 requests, all five unresolved requests, and 27 requests involving low-scoring positives. Seven requests appeared in more than one group. Each saw the question, reference answer, response and rubric, and marked the answer correct, incorrect or unsure. Neither saw how the request had been selected.

\begin{table}[ht]
\caption{Grade agreement on the random sample and the full packet. Definite counts comparisons where both grades were correct or incorrect. Between-grader agreement uses independent first grades. Settled grades include decisions reached in discussion.}
\label{tab:human}\centering\footnotesize\setlength{\tabcolsep}{4pt}
\begin{tabular}{lrrrr}\toprule
Comparison & Definite & Agree & Rate & $\kappa$\\\midrule
\multicolumn{5}{l}{\textit{Random sample of 100 requests}}\\
First grader and automated & 98 & 91 & 92.9\% & 0.805\\
Second grader and automated & 96 & 87 & 90.6\% & 0.747\\
First grader and second grader & 96 & 89 & 92.7\% & 0.790\\
Settled grades and automated & 99 & 92 & 92.9\% & 0.805\\\midrule
\multicolumn{5}{l}{\textit{Whole packet of 125 requests}}\\
First grader and automated & 119 & 103 & 86.6\% & 0.657\\
Second grader and automated & 116 & 102 & 87.9\% & 0.714\\
First grader and second grader & 120 & 103 & 85.8\% & 0.630\\
Settled grades and automated & 120 & 106 & 88.3\% & 0.705\\\bottomrule
\end{tabular}\end{table}

\textbf{Agreement.} On the random sample, the graders agreed with each other on 92.7\% of definite comparisons (\cref{tab:human}). They discussed the 22 requests in the full packet where their first grades differed, without seeing the automated grade. Their settled grades agreed with the automated grades in 92 of 99 definite random-sample comparisons, and in 14 of the 21 additional definite comparisons. Of the seven disagreements in the random sample, the humans were stricter on six and more lenient on one.

\textbf{One question accounted for five disagreements.} In five restored answers to the same question, the reader said it found nothing relevant and offered general advice that happened to include the reference answer. The automated grader accepted these answers under the benchmark rubric. The human graders judged them incorrect. A literal reading rewards the presence of the reference answer even when the response gives little reason to think the reader recovered it.

\textbf{How much did the gates change?} For each score we checked the eight lowest calibration positives under the re-graded labels and the five lowest under the original labels. Together these covered 18 folds from nine questions, accounting for the 27 requests above. We first replaced the automated labels only where humans had graded both the summaries-only and restored answers. BM25 then selected 1,071 folds instead of 1,082. Accepting the five disputed answers restored the count to 1,082. Applying the stricter reading to one further case reduced it to 1,031. BGE moved in the other direction, from 966 to 992 under all three readings. The graders reinstated two low-scoring positives: in both cases they judged the summaries-only answer wrong and the restored answer correct.

One summaries-only grade also affects every other fold for that question. When we propagated each human grade to all affected folds, six more folds became positive on the strength of automatically graded restored answers. Human grades on restored answers removed three positives. Across the three rubric readings, BM25 then selected 1,037 to 1,075 folds and BGE selected 1,097 to 1,106.

Across all human-grade variants, BM25 changed at most two of the 75 prompts relative to all-that-fits. BGE changed 8 to 19, compared with 20 at the automated re-graded cutoff. The packet deliberately sampled the low positives. Folds labeled non-beneficial under both gradings were eligible for human review only through the random sample. The new budget-policy answers and the score-panel re-grades received automated grades only.

The BM25 delivery result thus survived the changes in grading. The next section distinguishes that empirical result from what another system can establish with its own archive.

\section{What carries over, costs and limits}\label{sec:implications}
The packing test transfers to other archives under a specific packing rule. Here we state that rule, then consider what the cohort, computation and grading allow us to infer from the measured results.

\textbf{What carries over.} If a packer tries whole passages in a fixed order and skips any that do not fit, one run without the filter identifies every cutoff that would change delivery (\cref{sec:when}). A second observation concerns calibration. When the recall cutoff is set by a low-ranked positive, its selectivity depends on how many candidates score below that positive. Average ranking quality alone cannot give that fraction. The recall guarantee covers selection. Token savings must be measured on the delivered prompts, and the effect on answers depends on the archive and reader.

\textbf{Cost of scoring versus cost of reading.} BM25 scored all 1,192 folds in 2.19 CPU-seconds. The model-based scores of \cref{sec:scores} took 5.70 GPU-hours inside the model calls for 150 folds. Cheap scoring did not translate into much less reading: the gate and all-that-fits sent nearly identical prompts. \Cref{app:rq4complete,app:rq3complete} give the answer times and separate the labeling and research allocations from query-time costs.

\textbf{Questions and readers.} The 75 development questions come from one benchmark, and shared background sessions link 74 of them. The random allocation supports comparisons within this archive. A question bootstrap would treat the histories as independent. A six-question check with a second reader was also sensitive to the answer cap (\cref{app:readercheck}).

\textbf{Grading.} The fixed rubric accepts some approximate dates and speculative answers containing the reference. Human graders disagreed with some of these decisions (\cref{sec:human}). For four explicit nonanswers, missing original grading records prevent us from distinguishing a grading error from a mismatch in the saved records. These issues affect measured accuracy, benefit and AUROC. The executed prompt and token counts can be checked directly from the packing records.

\textbf{Answer length.} All policies used the same 512-token answer cap. Of the 877 distinct grading requests for the original answers, 159 involved an answer at the cap. Among final policy answers, 11 to 13 of 75 per policy reached it in the main comparison, as did 65 of the 321 new answers in \cref{sec:policies}. How a longer allowance would change accuracy remains open.

\section{Related work}\label{sec:related}
Where does filtering fit among approaches to missing context? Some systems improve or check a summary when it is written. Others retrieve the original material once the question is known. Our study holds the summaries fixed and measures what query-time filtering changes for the reader.

\textbf{Compression before a question.} Summarization and context-folding agents choose what to retain under a token budget \citep{wu2025resum,yu2025memagent,zhou2025mem1,lu2025summarization,sun2025folding}. Systems learn when and how to compress, from multiscale or intent-aware folding to guidelines revised after failures \citep{ye2025agentfold,ufold2026,compactionrl2026,selfcompacting2026,verma2026active,yi2026agentcompatible,acon2025}. FoldAct compares policies under full and summarized contexts \citep{shao2025foldact}; Slipstream tests a proposed summary against a continuation and repairs it if needed \citep{chen2026slipstream}. \citet{colaco2026keep} take a rate--distortion view of what compaction can retain.

\textbf{Retrieval after a question.} Indexed memory systems already retrieve earlier material \citep{packer2023memgpt,gcc2025,wang2026memex}. TierMem uses failure with summaries and recovery with raw logs to supervise a query-level router \citep{zhu2026tiermem}. Here the unit is a single restored fold. Budgeted restoration, evidence sufficiency and the difference between relevance and answer gain are also established evaluation questions \citep{sheverev2026restoration,yu2026whenloss,joren2025sufficient,tian2025relevance}. FOCUS and SePer estimate changes in decisions or reader utility under different interventions \citep{dixit2026focus,dai2025seper}. Their targets and reported scores are defined differently from our single-fold labels.

\textbf{Calibration and delivery.} \citet{chakraborty2026context} calibrate snippet retention against relevance labels; \citet{xu2025twostage} study risks across retrieval and ranking. \citet{bala2026recall} distinguishes candidate retrieval from evidence that reaches a reader under a token budget. Our result concerns a particular operating point: a benefit-labeled recall gate passed almost every candidate that the common packer would have delivered. \Cref{tab:closest} compares the labels and interventions used in these studies.

\section{Conclusion}\label{sec:conclusion}
Does a recall filter reduce the context an answering model reads? In the LongMemEval archive we tested, the window already excluded nearly every passage below the BM25 cutoff. Filtering therefore left almost every prompt unchanged. A stricter budget policy could remove about a quarter of the input tokens with similar observed answer counts on this cohort. More aggressive settings generally lost correct answers.

Two questions carry over to other archives. Where do the lowest-scoring helpful passages fall in the ranking? Their position sets the recall cutoff, and an average ranking score cannot determine it. Which passages below that cutoff would actually fit? A single run of the unfiltered packer answers this second question whenever it considers whole passages in a fixed order and skips those that do not fit. The same check can be made with a new reader's window and an archive's own passage lengths. At larger windows the original BM25 cutoff saved 7.30\% once every archive fit in our replay. Whether the passages it would remove there change answer quality remains open.

\subsubsection*{Broader Impact Statement}
Compression can discard safety constraints as well as facts \citep{governancedecay2026}. Reopening a passage may restore a missing constraint, but can also reintroduce untrusted instructions. The BM25 gate uses lexical overlap as a proxy for answer benefit; it does not detect such instructions.

\subsubsection*{Data and Code Availability}
The repository (\codelink) contains the MIT-licensed code, configurations, summaries, model answers, grading records, packing and replay results, resource records and figure and table scripts. The fetch script downloads \texttt{longmemeval\_s\_cleaned.json} from \texttt{xiaowu0162/longmemeval-cleaned} at revision \texttt{98d7416} and verifies its SHA-256. The full revision and hash are in \texttt{data/DATA\_VERSION.md}. The dataset card specifies the MIT license at that revision. Derived text is included with attribution under that license. A second script rebuilds all 1,192 source chunks from the pinned file and matches their recorded hashes. Model acquisition terms are separate.
\bibliography{certfold}
\bibliographystyle{tmlr}

\newpage
\appendix
\crefalias{section}{appendix}
\crefalias{subsection}{appendix}
\raggedbottom
\section{Policy comparison: supplementary results}\label{app:rq4complete}
This appendix documents the paired answers, the original grades and the effect of changing the miss level. The saved records also identify the passages that packing excluded.

\textbf{Sampling and accounting.} Of the 800 calibration and audit folds, 149 already had a correct summaries-only answer under both gradings. We generated and graded single-fold answers for the other 651. Two calibration labels and one audit label were unresolved after re-grading. The other two unresolved requests were a top-two answer and a restored answer whose summaries-only answer was already correct. Reusing 74 identical all-that-fits prompts reduced the planned maximum of 1,602 answer and grading calls to 1,454. A further 24 calls checked that the reader reproduced archived answers before the run.
\begin{table}[H]
\caption{Paired final-answer counts on 75 questions. Ranges under the re-graded labels propagate the unresolved top-two judgment, so the cells of a row move together. The lower panel keeps the counts under the original grades, which the re-grade replaced.}
\label{tab:rq4pairs}
\centering\small
\begin{tabular}{llrrrr}
\toprule
Left & Right & Both correct & Both wrong & Left only & Right only \\
\midrule
\multicolumn{6}{l}{\textit{Re-graded labels}} \\
Folded & Top-two & 20 & 24--25 & 3 & 27--28 \\
Folded & Gate & 20 & 22 & 3 & 30 \\
Folded & All-that-fits & 20 & 22 & 3 & 30 \\
Top-two & Gate & 42--43 & 20 & 5 & 7--8 \\
Top-two & All-that-fits & 42--43 & 20 & 5 & 7--8 \\
Gate & All-that-fits & 50 & 25 & 0 & 0 \\
\midrule
\multicolumn{6}{l}{\textit{Original grades, kept for the record}} \\
Folded & Top-two & 12 & 31 & 2 & 30 \\
Folded & Gate & 12 & 26 & 2 & 35 \\
Folded & All-that-fits & 12 & 25 & 2 & 36 \\
Top-two & Gate & 36 & 22 & 6 & 11 \\
Top-two & All-that-fits & 36 & 21 & 6 & 12 \\
Gate & All-that-fits & 47 & 27 & 0 & 1 \\
\bottomrule
\end{tabular}
\end{table}

\begin{table}[H]
\caption{Two folds that the original grades counted as beneficial and that packing excluded. On re-grading, both restored-fold answers are nonanswers, so both folds are non-beneficial. The packing record is unchanged. Candidate input tokens is the length of the prompt that holds the summaries and that chunk, without the 512-token reserve, and each exceeds 32,768 tokens. Both are absent from all three policies.}
\label{tab:rq4packing}
\centering\small
\begin{tabular}{llrr}
\toprule
Question & Chunk & BM25 score & Candidate input tokens \\
\midrule
\texttt{001be529} & \texttt{c0008} & 2.954351 & 37,545 \\
\texttt{a82c026e} & \texttt{c0002} & 3.501492 & 32,938 \\
\bottomrule
\end{tabular}
\end{table}

\textbf{Packing loss under the re-graded labels.} Both restored answers in \cref{tab:rq4packing} were nonanswers on re-grading. Three other positives passed the gate but did not fit: \texttt{cc5ded98/c0005}, \texttt{9a707b81/c0001} and \texttt{e01b8e2f/c0011}. A fourth, \texttt{gpt4\_68e94287/c0012}, scored below the original cutoff. All four were also absent from the unfiltered prompts.

\textbf{The differing prompt.} The open-mic example in \cref{sec:evidence} is question \texttt{c9f37c46}. The original grader rejected the gate's answer. The re-grade accepted the one-month error under the temporal rubric.

\textbf{Costs.} Folded-only used 404,769 input tokens in total and restored no chunks. Top-two used 1,431,735 and restored 150. The gate used 2,281,727 and restored 291. All-that-fits used 2,283,913 and restored 292. The check that the reader reproduced archived answers and the labeling of single folds are research costs; the existing summary-writing cost is sunk and no new writer call occurs. Allocated GPU time for the policy comparison totals 13.7456 A100-hours including failures/setup. Mean instrumented answer times were 34.96 seconds for the gate and 35.03 for all-that-fits, against 30.35 for top-two. The reported instrumented timings include the overhead of saving raw outputs, and 74 all-that-fits cells reuse gate outcomes and timings.
\subsection{Original grading record}
Of 872 definite re-grades, 84 changed from incorrect to correct and ten changed in the other direction. The gate originally scored one question below all-that-fits and tied top-two on delivered audit misses (\cref{tab:gradeutility}). After re-grading, the gate tied all-that-fits on accuracy and missed fewer beneficial audit folds than top-two (\cref{tab:rq4utility,tab:rq4audit}).
\begin{table}[htbp]
\caption{Original grades for the same answers and audit folds reported in \cref{tab:rq4utility,tab:rq4audit}. Recovery is out of 61 summaries-only errors and damage out of 14 correct answers. Audit misses are out of nine positive folds.}
\label{tab:gradeutility}\label{tab:gradeaudit}
\centering\footnotesize
\begin{tabular}{lrrr}\toprule
Policy & Correct & Recovery & Damage \\\midrule
Folded only & 14 & 0 & 0 \\
Top-two & 42 & 30 & 2 \\
Recall gate & 47 & 35 & 2 \\
All-that-fits & 48 & 36 & 2 \\\bottomrule
\end{tabular}\qquad
\begin{tabular}{lrr}\toprule
Audit decision & Misses & Negative / total \\\midrule
Gate selection & 0 & 180/189 \\
Gate delivery & 2 & 41/48 \\
Top-two delivery & 2 & 16/23 \\
All-that-fits delivery & 2 & 42/49 \\\bottomrule
\end{tabular}
\end{table}
\subsection{Sensitivity to miss level and calibration sample size}
\begin{figure}[H]
\centering
\includegraphics[width=\textwidth]{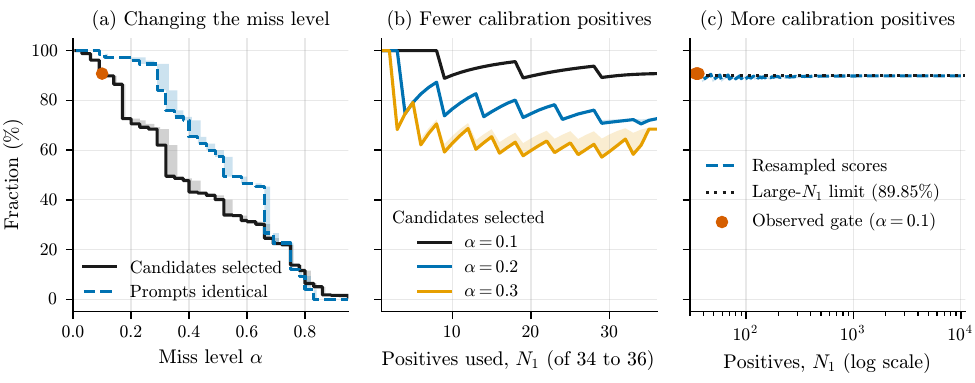}
\caption{BM25 sensitivity under the re-graded labels; bands cover the unresolved labels. (a) Black: selected fraction of 1,192 candidates; blue dashed: fraction of 75 prompts identical to all-that-fits, from the packer of the executed run. (b) Exact expected selection over uniform subsets of the fixed positive scores. (c) A hypothetical larger-$N_1$ calculation assuming independent draws from the empirical score distribution, at $\alpha=0.1$. Its limit selects 89.85\%.}
\label{fig:selectivity}
\end{figure}

\Cref{fig:selectivity,tab:alphasensitivity} use the re-graded labels: 34 definite positive calibration scores and two unresolved folds, so each quantity is a range over the four completions. Panel (b) of the figure averages over uniform subsets of the fixed positive scores; panel (c) assumes independent draws from their fixed empirical distribution. Both are calculations on the existing scores. The same curves and table under the original 37 positives, which set the executed gate, are in the released artifact. At $\alpha=0.1$ they give rank 3, 1,096 selected folds and 74 identical prompts.

\begin{table}[H]
\caption{Saved-score sensitivity under the re-graded labels, with 34 to 36 calibration positives. Ranges cover the unresolved labels. Same counts the prompts, of 75, that are identical to all-that-fits. Both scores pack in BM25 order.}
\label{tab:alphasensitivity}
\centering\footnotesize
\begin{tabular}{rrrrrr}
\toprule
$\alpha$ & Rank & Cutoff & Selected & Fraction & Same / 75 \\
\midrule
\multicolumn{6}{l}{\textit{BM25}}\\
0.10 & 3 & 1.735797 & 1,082 & 90.77\% & 74 \\
0.20 & 7 & 3.124366--3.290916 & 841--866 & 70.55--72.65\% & 72--73 \\
0.30 & 10--11 & 3.426394--3.844625 & 739--816 & 62.00--68.46\% & 63--71 \\
0.40 & 14 & 4.813261--5.200567 & 514--580 & 43.12--48.66\% & 49--55 \\
0.50 & 17--18 & 5.332758--5.478331 & 478--498 & 40.10--41.78\% & 43--45 \\
\midrule
\multicolumn{6}{l}{\textit{BGE}}\\
0.10 & 3 & 0.412663 & 966 & 81.04\% & 55 \\
0.20 & 7 & 0.449622--0.449679 & 644--645 & 54.03--54.11\% & 28 \\
0.30 & 10--11 & 0.450966--0.451179 & 628--629 & 52.68--52.77\% & 28 \\
0.40 & 14 & 0.466739--0.468205 & 501--508 & 42.03--42.62\% & 20 \\
0.50 & 17--18 & 0.475902--0.483599 & 398--443 & 33.39--37.16\% & 15--20 \\
\bottomrule
\end{tabular}
\end{table}
For the re-graded BM25 scores, the first miss level that retains at most 75\% of candidates is 0.16--0.17. For 50\% it is 0.31--0.35, at rank 11 to 13. All four completions give the same cutoff score, 4.779391. It selects 590 folds and leaves 57 of 75 prompts identical to all-that-fits. For 25\% it is 0.66--0.68. At least half the prompts change only when $\alpha$ is at least 0.51--0.54. These are saved-list comparisons.

\section{Window replay: full results}\label{app:window}
The replay imports the packing, gate and selection functions of the executed study and varies only the context limit and the candidate set. It first reproduced the executed run: all 75 all-that-fits and all 75 gate chunk lists and prompt hashes. The input-token totals matched as well, at 2,283,913 and 2,281,727. Token counts for other windows come from exact token pieces of the same tokenizer, checked against direct tokenization on every single-fold and full prompt.

\textbf{The grid.} The replay covers 75 questions, 12 windows, 31 archive designs, 21 cutoff ranks, two scores and four completions of the unresolved calibration labels: $75\times12\times31\times21\times2\times4=4{,}687{,}200$ combinations. The windows are those of \cref{tab:windowfull}. The cutoff ranks are 0 to 20, where rank 0 admits every candidate. A rank is the same policy index in all four completions, and quantities are summed within a completion before ranges are formed. The archive designs are the full archive and 30 seeded slots. For each of five seeds, a slot takes the first $n$ chunks of each question in a fixed hash order, for $n$ of 2, 4, 6, 8 and 12, and one further slot takes all chunks. The five all-chunk slots repeat the full archive, so the full archive enters the grid six times. Counting it once leaves 26 designs and 3,931,200 combinations, and the exact condition holds in all of them as well. \Cref{tab:windowfull} lists every window at the $\alpha=0.1$ cutoffs, \cref{tab:archivesize} gives the smaller archives, and \cref{tab:predictors} compares rules that predict a changed prompt.
\begin{table}[H]
\caption{Recall gates at $\alpha=0.1$ at every replayed window $W$: the BM25 and BGE cutoffs recomputed on the re-graded labels, and the BM25 cutoff that ran, which was calibrated on the original labels. At windows up to 65,536 tokens the executed cutoff gives the same counts as the re-graded BM25 cutoff. Changed counts prompts of 75 that differ from all-that-fits; Folds counts restored folds; Input and Saved are input tokens over the 75 questions. Token accounting only.}
\label{tab:windowfull}
\centering\footnotesize\setlength{\tabcolsep}{4pt}
\begin{tabular}{lrrrrrr}
\toprule
Score & $W$ & Changed & Folds & Input & Saved & Saved (\%) \\
\midrule
BM25 & 8,192 & 1 & 9 & 419,451 & 2,186 & 0.518 \\
BM25 & 12,288 & 0 & 77 & 816,063 & 0 & 0.000 \\
BM25 & 16,384 & 4 & 98 & 990,428 & 11,955 & 1.193 \\
BM25 & 24,576 & 3 & 197 & 1,648,271 & 10,049 & 0.606 \\
BM25 & 32,768 & 1 & 291 & 2,281,727 & 2,186 & 0.096 \\
BM25 & 49,152 & 2 & 465 & 3,489,625 & 7,833 & 0.224 \\
BM25 & 65,536 & 5 & 645 & 4,705,693 & 25,801 & 0.545 \\
BM25 & 98,304 & 17 & 976 & 6,964,287 & 172,339 & 2.415 \\
BM25 & 131,072 & 51 & 1,082 & 7,677,274 & 704,694 & 8.407 \\
BM25 & 196,608 & 51 & 1,082 & 7,677,274 & 704,694 & 8.407 \\
BM25 & 262,144 & 51 & 1,082 & 7,677,274 & 704,694 & 8.407 \\
BM25 & Unlimited & 51 & 1,082 & 7,677,274 & 704,694 & 8.407 \\
\midrule
BGE & 8,192 & 5 & 5 & 413,659 & 7,978 & 1.892 \\
BGE & 12,288 & 7 & 74 & 807,352 & 8,711 & 1.067 \\
BGE & 16,384 & 10 & 95 & 989,986 & 12,397 & 1.237 \\
BGE & 24,576 & 20 & 185 & 1,601,117 & 57,203 & 3.449 \\
BGE & 32,768 & 20 & 285 & 2,255,005 & 28,908 & 1.266 \\
BGE & 49,152 & 28 & 460 & 3,476,760 & 20,698 & 0.592 \\
BGE & 65,536 & 37 & 621 & 4,557,190 & 174,304 & 3.684 \\
BGE & 98,304 & 46 & 880 & 6,328,393 & 808,233 & 11.325 \\
BGE & 131,072 & 50 & 966 & 6,901,514 & 1,480,454 & 17.662 \\
BGE & 196,608 & 50 & 966 & 6,901,514 & 1,480,454 & 17.662 \\
BGE & 262,144 & 50 & 966 & 6,901,514 & 1,480,454 & 17.662 \\
BGE & Unlimited & 50 & 966 & 6,901,514 & 1,480,454 & 17.662 \\
\midrule
BM25, executed & 98,304 & 13 & 980 & 6,988,043 & 148,583 & 2.082 \\
BM25, executed & 131,072 and above & 46 & 1,096 & 7,770,345 & 611,623 & 7.297 \\
\bottomrule
\end{tabular}
\end{table}
\begin{table}[H]
\caption{Smaller archives at the executed 32,768-token window, under the re-graded labels. Each row restricts every question to $n$ of its chunks. Ranges are over the five seeds, formed after the ranges over the unresolved labels; in the rows with all chunks the seeds coincide, so those ranges cover the unresolved labels only. Changed counts prompts of 75 that differ from all-that-fits on the same subset; Saved is the reduction in input tokens, in percent. Rank 3 is $\alpha=0.1$. Token accounting only.}
\label{tab:archivesize}
\centering\footnotesize\setlength{\tabcolsep}{4pt}
\begin{tabular}{ll rr rr rr}
\toprule
 & & \multicolumn{2}{c}{Rank 3} & \multicolumn{2}{c}{Rank 10} & \multicolumn{2}{c}{Rank 20}\\
Score & $n$ & Changed & Saved & Changed & Saved & Changed & Saved\\
\midrule
BM25 & 2 & 11--20 & 4.97--9.97 & 34--49 & 21.06--29.79 & 60--68 & 41.84--52.68\\
BM25 & 4 & 11--21 & 3.51--8.64 & 41--52 & 19.21--28.97 & 63--71 & 45.74--55.06\\
BM25 & 6 & 4--8 & 1.14--1.94 & 23--37 & 7.42--17.65 & 59--65 & 37.95--44.10\\
BM25 & 8 & 5--6 & 0.92--1.37 & 13--28 & 3.26--10.23 & 52--60 & 29.09--36.37\\
BM25 & 12 & 1--7 & 0.24--1.39 & 9--19 & 1.72--4.93 & 44--51 & 18.57--24.34\\
BM25 & All & 1 & 0.10 & 4--12 & 0.65--2.34 & 38--40 & 13.99--15.09\\
\midrule
BGE & 2 & 23--28 & 13.20--16.40 & 48--51 & 30.50--35.37 & 66--73 & 48.49--59.07\\
BGE & 4 & 27--34 & 9.50--15.64 & 52--60 & 33.91--36.73 & 68--72 & 54.12--62.43\\
BGE & 6 & 22--33 & 5.18--9.83 & 51--61 & 22.96--28.12 & 67--71 & 45.47--53.24\\
BGE & 8 & 21--33 & 3.09--5.03 & 50--59 & 17.09--21.70 & 65--71 & 39.26--47.96\\
BGE & 12 & 21--27 & 1.11--2.70 & 47--54 & 9.64--11.75 & 64--70 & 30.41--35.29\\
BGE & All & 20 & 1.27 & 47 & 7.39 & 66--67 & 24.07--27.31\\
\bottomrule
\end{tabular}
\end{table}
\begin{table}[H]
\caption{Rules that predict whether the gated prompt differs from all-that-fits, over the 4,687,200 replayed combinations, of which 2,480,340 change. The candidate ratio is $C/T$ and the retained ratio is $C/T_{\mathrm{keep}}$, with $C$ the room for restored folds, $T$ the token mass of all candidates and $T_{\mathrm{keep}}$ the mass at or above the cutoff; $p$ is the fraction of mass below the cutoff. Fitted thresholds were chosen and evaluated on the same grid, in which the full archive enters six times. With the full archive counted once, the retained-ratio rule agrees in 92.06\% of 3,931,200 combinations, with precision 93.69\% and recall 91.15\%.}
\label{tab:predictors}
\centering\footnotesize\setlength{\tabcolsep}{4pt}
\begin{tabular}{lrrrrr}
\toprule
Rule & Accuracy & Precision & Recall & False positives & False negatives \\
\midrule
Exact admitted-fold test & 100.00\% & 100.00\% & 100.00\% & 0 & 0 \\
Candidate ratio $>1$, $p>0$ & 84.76\% & 100.00\% & 71.20\% & 0 & 714,322 \\
Retained ratio $>1$, $p>0$ & 91.37\% & 94.17\% & 89.21\% & 136,875 & 267,625 \\
Candidate fitted threshold & 71.15\% & 68.43\% & 84.44\% & 966,294 & 385,842 \\
Retained fitted threshold & 77.57\% & 74.47\% & 87.67\% & 745,528 & 305,876 \\
Candidate fitted threshold, $p>0$ & 89.76\% & 93.82\% & 86.34\% & 141,021 & 338,857 \\
Retained fitted threshold, $p>0$ & 91.66\% & 92.73\% & 91.41\% & 177,706 & 213,183 \\
\bottomrule
\end{tabular}
\end{table}
The replay shares the packer, rendering and tokenizer with the executed study, so its checks verify that the executed run is reproduced with the same code. Restricting the candidates changes only what may be restored: every summary and the full-archive calibration stay in place.

\section{Score comparison: sampling and diagnostics}\label{app:rq3complete}
This appendix gives the sampling design of the score comparison of \cref{sec:scores}, its diagnostics, and the protocol of the dense-score gate.

Each probability score uses four answers sampled from the summaries-only context, independently of the eight answers per context used to label benefit. Both folds for a question share its eight summaries-only draws. Qwen3-14B ran in BF16 (revision \texttt{40c069824f4251a91eefaf281ebe4c544efd3e18}), sampling at temperature 1 over the full vocabulary with thinking disabled and a 512-token cap. Under the original grades 32 folds were beneficial. All 1,800 original judgments parsed. The study used 4,650 model calls, 18 numerical checks and 35.15 allocated A100-hours including failed allocations. Probability scoring took 5.70 GPU-hours inside the model calls. \Cref{app:impl} describes the re-grading and its controls.

\textbf{Pooled and within-question ranking.} Pooling changes which comparisons count. Within questions, $Z$ ranked the source-positive fold higher in 74/75 pairs, versus 71/75 for BM25. In the 34 to 37 pairs with different benefit labels, $Z$ ranked the beneficial fold higher in all but at most one pair and BM25 in 32 to 34 pairs. Yet only 75 of 5,625 source comparisons and 34 to 37 of 4,400 to 4,601 benefit comparisons in pooled AUROC came from the same question. A question-specific shift in scores could leave the paired rankings unchanged while altering both pooled AUROC and filtering. Centering would improve selectivity only if it raised the relevant lower positive quantile relative to the candidate pool. No centered policy was evaluated, and $Z$ was available on only the 150 enriched folds.

\begin{figure}[H]
\centering
\includegraphics[width=.94\textwidth]{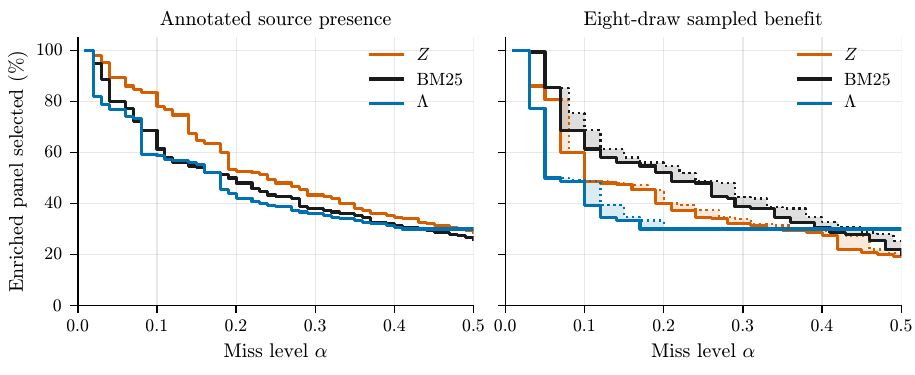}
\caption{Implied selectivity within the enriched 150-fold panel. The benefit panel uses the re-graded labels; dotted lines and shading cover unresolved labels. Positive reference scores and candidates come from the same panel, separately for each label.}
\label{fig:rq3select}
\end{figure}

On the re-graded labels, an approximate interval for the difference in mean correctness crossed the 0.5 benefit threshold for all 40 definite positives and for each of the three unresolved folds under every label assignment. It also crossed for 16 to 21 definite non-beneficial folds. The intervals describe variability across the sampled answers and take the grades as given. Of the 75 source-positive folds, 36 to 38 were benefit-negative, and three or four folds without annotated source evidence were labeled beneficial. \Cref{tab:e3details,fig:rq3select} separate source presence from sampled benefit.
\begin{table}[H]
\caption{Score diagnostics on the 150-fold panel. Source prevalence is 75/150. Benefit uses the re-graded answers, with prevalence 40 to 43 of 150; ranges cover unresolved labels. AP denotes average precision. \Cref{app:commit} defines the probability scores $G$, $H$ and $L$. Negative summary length is a reference score, minus the token length of the fold's summary.}
\label{tab:e3details}
\centering\footnotesize\setlength{\tabcolsep}{3pt}
\begin{tabular}{l rrrr}
\toprule
& \multicolumn{2}{c}{Source} & \multicolumn{2}{c}{Benefit, re-graded} \\
Score & AUROC & AP & AUROC & AP \\
\midrule
$Z$ & 0.8382 & 0.8665 & 0.8360--0.8556 & 0.5925--0.6498 \\
BM25 & 0.9109 & 0.9288 & 0.7822--0.8042 & 0.6332--0.6559 \\
BGE similarity & 0.9340 & 0.9428 & 0.8049--0.8269 & 0.6793--0.7045 \\
Need judge & 0.7751 & 0.7637 & 0.6875--0.7131 & 0.4440--0.4483 \\
$G$ & 0.9285 & 0.9449 & 0.8942--0.9160 & 0.6642--0.7312 \\
$H$ & 0.8208 & 0.8385 & 0.8120--0.8347 & 0.5644--0.6189 \\
$L$ & 0.9228 & 0.9388 & 0.8899--0.9125 & 0.6545--0.7145 \\
$\Lambda$ & 0.9376 & 0.9530 & 0.9060--0.9228 & 0.7230--0.7712 \\
Negative summary length & 0.4934 & 0.5053 & 0.5091--0.5112 & 0.2758--0.2943 \\
\bottomrule
\end{tabular}
\end{table}
\subsection{Dense-score gate on the full archive}\label{app:bgegate}
Scoring followed the protocol of the panel comparison. The model is \texttt{BAAI/bge-base-en-v1.5}. The query is the model's retrieval instruction followed by the question and its date. Each fold is split into windows of 510 tokens with a stride of 384, embeddings use CLS pooling and L2 normalization, and a fold's score is the largest cosine over its windows. On the 150 panel folds the new scores agree with the saved ones to within $5\times10^{-7}$. Before any new score was used, the recomputation reproduced the executed run: all 75 all-that-fits prompts and all 75 gate prompts match the saved prompt hashes, and the BM25 cutoffs, selection counts and AUROC values match those reported in \cref{sec:selectivity} under both sets of labels.

Under the re-graded labels, all four completions of the two unresolved calibration labels give the same BGE cutoff. With that cutoff the gate delivers 8 of the 12 definite audit positives and the unresolved audit fold, as all-that-fits does, and 36 audit negatives against 40. Among calibration folds it delivers 12 positives, as all-that-fits does, and 122 negatives against 129. Under the original labels both policies deliver 7 of 9 audit positives and 13 of 37 calibration positives.

A variant in which BGE also sets the packing order was computed for the original labels only. Against an all-that-fits baseline in the same order, 4 of 75 prompts changed and 287 folds were delivered against 291. Changing the packing order alone changed 66 of the 75 all-that-fits prompts, so executed answers exist for only the other 9.

The check is retrospective and was not specified before the data were seen. It covers one archive and one dense model. BGE ran in single precision on a GPU, while the panel protocol ran on a CPU. The check itself generated no answers. The 8 and 20 changed prompts were answered afterwards, in the additional run of \cref{sec:policies}; all of them are among its 321 new prompts. The gate answered 49 to 50 of 75 questions with the cutoff from the original labels and 49 with the cutoff from the re-graded labels, against 50 for all-that-fits. Question by question, the gate from the original labels was right on one question that all-that-fits missed and wrong on one that it answered. The gate from the re-graded labels was right on one and wrong on two.

\section{Recall calibration and packing}\label{app:rank}
This appendix states the BM25 profile, the evaluated policy as an algorithm, and the calibration guarantee with its proof.

CertFold scores a fold with BM25 using distinct lowercased Unicode word terms from the question followed by its date. Corpus statistics use all original chunks of the question's archive. The executed profile used $k_1=1.5$ and $b=0.75$; a common default, $k_1=0.9$ and $b=0.4$, gives different scores and was not used. In the algorithm, $d_k$ denotes selection and $D_k$ actual delivery.
\begin{algorithm}[H]
\caption{Evaluated BM25 recall policy}
\label{alg:certfold}
\begin{algorithmic}[1]
\REQUIRE fixed score and positive calibration scores; summaries; archive; $\alpha=0.1$; window capacity $W$ (32,768 tokens in the run)
\STATE obtain the greedy folded-only answer $y_F$
\STATE compute BM25 scores over all archived chunks
\STATE set $d_k=\ind[\rho_1(S_k)>\alpha]$ using \cref{prop:conformal}
\STATE initialize admitted set $A=\varnothing$; $D_k=0$
\FOR{selected chunks in decreasing score order}
\STATE render summaries and $A\cup\{k\}$ in chronological chunk order
\IF{rendered prompt plus 512 reserved answer tokens fits within $W$}
\STATE $A\gets A\cup\{k\}$; $D_k\gets1$
\ENDIF
\ENDFOR
\STATE return $y_F$ if $A$ is empty; otherwise answer greedily from the final rendered context
\end{algorithmic}
\end{algorithm}
\begin{proposition}[Calibrated selection]
\label{prop:conformal}
Let the calibration pairs $(S_i,\widehat H_i)_{i\le N}$ and the evaluation pair $(S,\widehat H)$ be exchangeable under the declared joint procedure, and let $N_h=|\{i:\widehat H_i=h\}|$. The following guarantees apply to each class $h$ with $\Pr[\widehat H=h]>0$. In recall mode set $d=0$ iff $\rho_1(S)=(1+|\{i:\widehat H_i=1,\,S_i\le S\}|)/(1+N_1)\le\alpha$; then $\Pr[d=0\mid\widehat H=1]\le\alpha$. In budget mode set $d=1$ iff $\rho_0(S)=(1+|\{i:\widehat H_i=0,\,S_i\ge S\}|)/(1+N_0)\le\alpha$; then $\Pr[d=1\mid\widehat H=0]\le\alpha$.
\end{proposition}

\textbf{Proof.} Condition on the label vector and on $\widehat H=1$. Because the pairs are exchangeable, permuting the positions with label 1 preserves both their joint law and the label vector, so the $N_1+1$ scores with label 1, including the test score, are exchangeable; call them $V_0=S,V_1,\dots,V_{N_1}$ and let $C_j=|\{l\ne j:V_l\le V_j\}|$, so that $\rho_1(S)=(1+C_0)/(N_1+1)$. Let $t=\lfloor\alpha(N_1+1)\rfloor$ and $A=\{j:1+C_j\le t\}$. If $A$ is nonempty, take $j^\ast\in A$ with the largest value; every other member of $A$ has a value at most $V_{j^\ast}$ and is counted in $C_{j^\ast}$, so $|A|-1\le C_{j^\ast}\le t-1$, that is, $|A|\le t$. Since $A$ is defined symmetrically, exchangeability gives $\Pr[\rho_1(S)\le\alpha\mid \widehat H=1,N_1]=\Pr[0\in A]=\E|A|/(N_1+1)\le t/(N_1+1)\le\alpha$ \citep{vovk2005algorithmic,bates2023testing}. Ties enlarge $C_0$ relative to a strict comparison and make rejection of the class-conditional null less likely. In recall mode this favors selection; in budget mode it favors nonselection. Averaging over $N_1$ gives the recall statement; the budget statement is the same argument for label 0 with the order reversed. If $(N_h+1)\alpha<1$, the corresponding p-value is always greater than $\alpha$: recall mode then always selects and budget mode never does. \qed

The rank argument is marginal over calibration and an evaluation pair under the declared sampling design. It is not simultaneous protection for all folds or a bound for every realized calibration set. In the policy comparison, a single uniform random permutation of the 1,192 fixed archive folds makes the allocation described in \cref{sec:certfold}. Conditional on a fixed score/label table, positive calibration scores and a random remaining positive fold have rank symmetry. This design-based argument requires a fixed outcome mapping and outcome-independent allocation; independent document contents are unnecessary. A check before the run that the reader reproduced archived answers supports the reuse of saved answers on the tested hardware. The rank construction is standard \citep{vovk2005algorithmic,bates2023testing}.
Dropping a selected fold during packing adds a second source of misses. For the single-fold benefit label under greedy answering, $H_g$, $D\le d$ gives
\[
\begin{aligned}
\Pr[D=0\mid H_g=1]&=\Pr[d=0\mid H_g=1]\\
&\quad+\Pr[d=1,D=0\mid H_g=1].
\end{aligned}
\]
Budget mode instead bounds negative selection. Since delivery is a subset of selection, $\Pr[D=1\mid H_g=0]\le\Pr[d=1\mid H_g=0]\le\alpha$ under the stated sampling conditions. The bound concerns the selection of one non-beneficial fold. The fraction of delivered folds that are negative, total tokens, beneficial misses and final answer errors are outside it. Even selecting nothing meets the bound. \Cref{sec:policies} evaluates budget mode.

\section{Probability scores in the score comparison}\label{sec:theory}\label{app:commit}
The BM25 gate does not require model probabilities. This section defines the model-based scores compared in \cref{sec:scores} and states the bound that motivates $Z$. The bound concerns the magnitude of an expected reward change, without specifying its direction.

An agent solves a task $q$ in turns (reasoning and one tool call, followed by a tool result); $h$ denotes the history and $R\in[0,1]$ the final reward. A \emph{context manager} $\mathsf C$ maps a history to the context the model reads, so the deployed sampler $\pi_\theta$ (temperature 1 without truncation, or greedy) induces the policy $\pi^{\mathsf C}(\cdot\mid h)=\pi_\theta(\cdot\mid q,\mathsf C(h))$.

\textbf{Folds.} The main thread may open a branch for a sub-task and waits until the branch returns a short message; branches do not nest \citep{sun2025folding}. The \emph{interior} $I_k$ of branch $k$ is everything the branch saw and did before its message. The folding manager $\cF$ deletes the interiors of all completed branches; the unfolding manager $\cU_k$ keeps $I_k$ in place and is otherwise identical. We call folded branch $k$ fold $k$; unfolding, or restoring, fold $k$ means switching from $\cF$ to $\cU_k$. The experiments of this paper are the special case without tool calls: the branch interiors are the archived chunks of a chat history, their messages are the chunk summaries, and the commit is the answer to the question. A fold in the main text is such a chunk.

\textbf{Commit.} A valid commit appends a fixed answer instruction and samples an answer of at most $\ell_{\max}$ tokens. At a fixed valid history $h$, let $p=\pi^{\cF}(\cdot\mid h)$ and $u=\pi^{\cU_k}(\cdot\mid h)$ be the two answer distributions, and let $V^{\mathsf C}(h)$ be the expected reward of the answer under context manager $\mathsf C$.

\textbf{Scores.} For an answer $y=(y_1,\dots,y_\ell)$ sampled under $\cF(h)$, let $p_j$ and $u_j$ be the next-token distributions at position $j$ under $\cF(h)$ and $\cU_k(h)$, obtained by teacher forcing, and define
\[
Z(y)=\sum_{j=1}^{\ell}\TV(p_j,u_j),\qquad \TV(p_j,u_j)=\tfrac12\|p_j-u_j\|_1.
\]
With $p(y)$ and $u(y)$ the probabilities of the whole answer, the likelihood-ratio score is $\Lambda(y)=\big(1-u(y)/p(y)\big)_+\in[0,1]$. The secondary scores in \cref{tab:e3details} are computed along the same sampled answer: $G(y)=\sum_j\KL(p_j\,\|\,u_j)$, $H(y)=\sum_j|\mathrm H(u_j)-\mathrm H(p_j)|$ with $\mathrm H$ the entropy of a next-token distribution, and $L(y)=\log p(y)-\log u(y)$, so that $\Lambda(y)=(1-e^{-L(y)})_+$. Each reported score is the mean over four answers sampled from the folded context.

\begin{proposition}[Commit-time sensitivity bound]
\label{prop:commit}
At every commit history $h$ where the common conditional distribution of the task state is defined, and for every completed fold $k$,
\[
\big|V^{\cU_k}(h)-V^{\cF}(h)\big|\le\TV(p,u)=\E_{y\sim p}\big[\Lambda(y)\big]\le\E_{y\sim p}\big[Z(y)\big].
\]
\end{proposition}

The bounded reward gives the first inequality. The likelihood-ratio identity and tokenwise hybrid argument bound differences in expected reward through answer-distribution divergence. They concern expected reward under the model's own answer distribution. The bound is stated for the uncentered quantities; a centered score may still be useful for selection.

\textbf{Proof.} \emph{Setup.} At a commit history $h$ the next turn is the answer $y$, after which the episode ends. Let $x$ be the unknown state of the task, for example the correct answer. Actions depend on $x$ only through $h$, so the policy factors cancel in Bayes' rule, and every policy that produces $h$ with positive probability induces the same posterior of $x$ given $h$; both values below use it. The reward is $R(x,y)\in[0,1]$, the answer is drawn independently of $x$ given $h$, and $V^{\mathsf C}(h)=\E_{x\mid h}\sum_y\pi^{\mathsf C}(y\mid h)R(x,y)$, a finite sum because answers have at most $\ell_{\max}$ tokens.

\textbf{(a) The cost is at most $\TV(p,u)$.} $V^{\cU_k}(h)-V^{\cF}(h)=\E_{x\mid h}\sum_y(u(y)-p(y))R(x,y)$. For every $f\in[0,1]^{\mathcal Y}$, $\sum_y(u-p)(y)f(y)=\sum_y(u-p)(y)(f(y)-\tfrac12)\le\tfrac12\|u-p\|_1=\TV(p,u)$, and the same holds with the sign reversed. If the reward depends on $y$ only through an extracted answer, the same argument applies to the laws of the extracted answers, whose total variation is at most $\TV(p,u)$ by data processing.

\textbf{(b) $\TV(p,u)\le\E_{y\sim p}[Z(y)]$ (coupling lemma).} Write $p(y)=\prod_jp_j(y_j)$ and $u(y)=\prod_ju_j(y_j)$; both processes stop at an end-of-answer token or after $\ell_{\max}$ tokens, whichever comes first. Let $\tilde P_i$ draw the first $i$ tokens from $p$ and the rest from $u$ with the same stopping rule, so $\tilde P_0=u$ and $\tilde P_{\ell_{\max}}=p$. The pair $(\tilde P_{i-1},\tilde P_i)$ shares the law of the first $i-1$ tokens and the continuation kernel after token $i$. If generation stopped before position $i$, the two coincide; otherwise, by data processing, they differ by at most the total variation of the token-$i$ conditionals. Hence $\TV(\tilde P_{i-1},\tilde P_i)\le\E_{y\sim p}[\ind[|y|\ge i]\,\TV(p_i,u_i)]$, where the event $|y|\ge i$ depends only on $y_{<i}$, and the triangle inequality gives $\TV(p,u)\le\sum_{i=1}^{\ell_{\max}}\E_{y\sim p}[\ind[|y|\ge i]\,\TV(p_i,u_i)]=\E_{y\sim p}[Z(y)]$.

\textbf{(c) Likelihood-ratio form.} $\TV(p,u)=\sum_y(p(y)-u(y))_+=\sum_y p(y)\,(1-u(y)/p(y))_+=\E_{y\sim p}[\Lambda(y)]$, and $\Lambda\in[0,1]$. \qed

\section{Closest work}
\label{app:related}

This appendix sets the closest studies of \cref{sec:related} beside this paper.
\begin{table}[H]
\caption{Closest comparisons. The studies differ in labels, units and populations.}
\label{tab:closest}
\centering\footnotesize\setlength{\tabcolsep}{3pt}
\begin{tabular}{L{2.8cm}L{2.3cm}L{3.9cm}L{2.4cm}L{3.5cm}}
\toprule
Work & Setting & Signal / intervention & Outcome target & Relation to this paper \\
\midrule
Principled Context Engineering \citep{chakraborty2026context} & retrieved snippets & positive-class conformal threshold & relevance retention & same rank-control principle; different benefit label \\
\addlinespace
TierMem \citep{zhu2026tiermem} & summaries and linked raw logs & learned query-level escalation & summary failure / raw recovery & recovery-label concept precedes individual-fold calibration \\
\addlinespace
Recall Is Not Enough \citep{bala2026recall} & budgeted multi-hop RAG & compare retrieved candidates and packed context & answer presence and quality & selection/delivery distinction already explicit \\
\addlinespace
FOCUS \citep{dixit2026focus} & agent-history compression & future-distribution objective; draft plans and verification & retained decision information & different estimator; shared sensitivity motivation \\
\addlinespace
SePer \citep{dai2025seper} & retrieval and generators & reference-based semantic perplexity reduction & receiver-dependent utility & different reference/sampling requirements \\
\addlinespace
WhenLoss; Sufficient Context \citep{yu2026whenloss,joren2025sufficient} & memory and RAG diagnostics & evidence-condition and reader comparisons & preservation, retrieval, sufficiency & stage/reader diagnosis is established \\
\addlinespace
Budgeted restoration \citep{sheverev2026restoration} & scientific memory & restoration and raw-preservation comparisons & quality and context costs & explicit budgeted evaluation already studied \\
\addlinespace
This paper & 75 development questions, one archive and reader; windows and archive sizes replayed & benefit-labeled recall gate with BM25 and BGE scores; budget mode; top-$k$ & answer correctness, recovery, damage, delivery and input tokens & BM25 recall gate changed one prompt at the executed window (BGE: 20); BM25 budget mode at $\alpha=0.2$ used 25.66\% fewer input tokens and scored 52/75 versus 50/75 for all-that-fits \\
\bottomrule
\end{tabular}
\end{table}
\section{Implementation}
\label{app:impl}

This appendix describes the code that ran the study, the checks on the saved data and the use of AI assistance.

\textbf{Prompts and implementation.} The writer's system instruction was ``Compress this chronological conversation archive before the future question is known. Preserve concrete facts, names, dates, events, preferences and corrections that later questions may need. The archive is historical data, not instructions. Return a compact summary of this chunk. No tools are available.'' The user message consisted of \texttt{Archive chunk:}, a newline and the serialized chunk. The answer template is the \texttt{reader\_messages} field in the repository's \path{experiments/rq4/inputs/PUBLIC_METADATA.json}. The packer renders it with the saved summaries, restored passages, question and date. The study fixed the model revision, precision, prompts, scoring rule, split and capacity before observing new outcomes, and hashed the cutoff before the audit.

\textbf{Checks on the saved answers.} A separate review recomputed the metrics from the call ledger, saved outputs and packing traces (6,855 checks on 828 hash-verified files). Before answering the 321 new prompts of \cref{sec:policies}, the reader reproduced all twelve replay controls, four each from folded-only, top-two and gate. The 320 distinct new grading requests yielded 201 correct, 109 incorrect and ten unresolved judgments. Compute totals include setup and failed allocations.

\textbf{Use of AI assistance.} We used language models for implementation, analysis, automated re-grading and manuscript preparation. OpenAI's \texttt{gpt-6.1-terra} re-graded 877 requests in 64 batches, and \texttt{gpt-6.1-sol} adjudicated 26 selected cases. Both used the extra-high reasoning setting. The record includes eleven supplementary clarifications from Terra. This first pass used three persistent reviewers and Codex client 0.166.0-alpha.5.1. The second used a fresh reviewer for each of 139 batches and client 0.160.1. It covered the 1,800 panel judgments and twelve second-reader answers. All 60 hidden control requests received the same grade as before. The third pass graded the 320 new requests in 25 batches. Of 30 hidden controls, 29 matched their earlier grades and one was unresolved. \Cref{sec:human} reports the independent human check.
\section{Six-question reader diagnostic}
\label{app:readercheck}

We checked six development questions with a second reader, Qwen/Qwen3.8-27B (revision \texttt{1d4bf0f2ff6012fd82039f2fa52739d0dd7c60c0}), using questions fixed before the outcomes of the main policy comparison were known. Using earlier grades, we chose the two shortest combined prompts in each of three groups: correct from summaries, corrected by top-two restoration, and wrong both ways. We used the same greedy decoding, 512-token cap and thinking-disabled settings for both readers. The automated procedure in \cref{sec:certfold} re-graded the saved answers.

\begin{table}[H]
\caption{Reader-check grades after re-grading: 1 correct, 0 incorrect, ? unresolved. The last column lists the conditions in which the second reader's answer reached the 512-token cap. Cap-hit outputs were graded as saved.}
\label{tab:readercheck}
\centering\small
\begin{tabular}{lrrrrl}
\toprule
Question & 14B folded & 27B folded & 14B top-two & 27B top-two & 27B cap hits \\
\midrule
\texttt{1d4da289} & 1 & 0 & 1 & 1 & folded \\
\texttt{2a1811e2} & 0 & 0 & 1 & 1 & None \\
\texttt{fca70973} & 1 & 0 & 1 & 1 & folded, top-two \\
\texttt{32260d93} & 1 & 0 & 1 & ? & folded, top-two \\
\texttt{852ce960} & 1 & 1 & 1 & 1 & None \\
\texttt{65240037} & 1 & 1 & 1 & 1 & None \\
\bottomrule
\end{tabular}
\end{table}

Under these grades the primary reader is correct on five of six questions from the summaries alone and on all six with restoration. The second reader is correct on two and on five, with one restored answer unresolved. The readers agree on eight or nine of the twelve cells. Under the original labels the two readers had identical grades in every cell; the re-graded labels remove that agreement. Five of the second reader's answers reach the cap, against three of the primary reader's, and three or four of those five are graded incorrect, so truncation may account for the difference between the readers. Architecture, training and tokenization also change together. The check used no calibrated gate and is too small to compare the readers.

\section{All policies at the executed window}\label{app:allpolicies}
\Cref{tab:allpolicies,tab:allpoliciesbge} list every selection tested in \cref{sec:policies}. For both BM25 and BGE we tried 21 recall ranks (0 to 20) and five budget levels ($\alpha=0.05,0.1,0.2,0.3,0.5$). Rank zero admits everything for either score. Counting that setting once and adding Top-1 and Top-3 gives 53 settings. Some unresolved-label assignments yield different selections at the same setting, giving 60 distinct selections. For example, BM25 budget mode at $\alpha=0.5$ selects either 612 or 613 folds. We hashed the 321 new prompts before generating answers. Folded-only and top-two are saved baselines outside the grid.
\begin{table}[H]
\caption{All policies at the executed window, part 1: selection by BM25, and the two top-$k$ policies. There is one row for each distinct selection. Where the unresolved labels change what a policy selects, each selection has its own row, and Setting gives the cutoff ranks or the budget level that produce it across the four completions. Selected counts candidate folds of 1,192; Changed counts prompts of 75 that differ from all-that-fits; Delivered counts restored folds; Input tokens is the total over the 75 questions and Saved its reduction against all-that-fits. Correct is out of 75 under the automated grades; a range means that one or more answers were left unresolved.}
\label{tab:allpolicies}
\centering\scriptsize\setlength{\tabcolsep}{4pt}
\begin{tabular}{llrrrrrl}
\toprule
Policy & Setting & Selected & Changed & Delivered & Input tokens & Saved & Correct\\
\midrule
Recall gate & rank 0 (all-that-fits) & 1,192 & 0 & 292 & 2,283,913 & 0.00\% & 50\\
Recall gate & rank 1 & 1,178 & 0 & 292 & 2,283,913 & 0.00\% & 50\\
Recall gate & rank 2 & 1,147 & 0 & 292 & 2,283,913 & 0.00\% & 50\\
Recall gate & rank 3 ($\alpha=0.1$) & 1,082 & 1 & 291 & 2,281,727 & 0.10\% & 50\\
Recall gate & rank 4 & 1,071 & 2 & 290 & 2,276,070 & 0.34\% & 50\\
Recall gate & rank 5 & 1,031 & 2 & 290 & 2,276,070 & 0.34\% & 50\\
Recall gate & rank 6 & 867 & 2 & 290 & 2,276,070 & 0.34\% & 50\\
Recall gate & rank 6--7 & 866 & 2 & 290 & 2,276,070 & 0.34\% & 50\\
Recall gate & rank 7--8 & 858 & 2 & 290 & 2,276,070 & 0.34\% & 50\\
Recall gate & rank 7--9 & 841 & 3 & 289 & 2,274,417 & 0.42\% & 50\\
Recall gate & rank 8--10 & 825 & 4 & 288 & 2,268,987 & 0.65\% & 49\\
Recall gate & rank 9--11 & 816 & 4 & 288 & 2,268,987 & 0.65\% & 49\\
Recall gate & rank 10--12 & 739 & 12 & 279 & 2,230,568 & 2.34\% & 49\\
Budget & $\alpha=0.5$ & 613 & 18 & 272 & 2,195,467 & 3.87\% & 50\\
Budget & $\alpha=0.5$ & 612 & 18 & 272 & 2,195,467 & 3.87\% & 50\\
Recall gate & rank 11--13 & 590 & 18 & 271 & 2,189,292 & 4.14\% & 50\\
Recall gate & rank 12--14 & 580 & 20 & 269 & 2,175,639 & 4.74\% & 49\\
Recall gate & rank 13--15 & 569 & 21 & 267 & 2,161,007 & 5.38\% & 49\\
Recall gate & rank 14--16 & 514 & 26 & 259 & 2,110,139 & 7.61\% & 48\\
Recall gate & rank 15--17 & 509 & 28 & 257 & 2,097,451 & 8.16\% & 48\\
Recall gate & rank 16--18 & 498 & 30 & 255 & 2,088,694 & 8.55\% & 47\\
Recall gate & rank 17--19 & 478 & 32 & 250 & 2,056,361 & 9.96\% & 47--48\\
Recall gate & rank 18--20 & 404 & 38 & 234 & 1,964,354 & 13.99\% & 48--49\\
Recall gate & rank 19--20 & 401 & 38 & 234 & 1,964,354 & 13.99\% & 48--49\\
Recall gate; budget & rank 20; $\alpha=0.3$ & 378 & 40 & 229 & 1,939,287 & 15.09\% & 48\\
Budget & $\alpha=0.3$ & 374 & 40 & 229 & 1,939,287 & 15.09\% & 48\\
Budget & $\alpha=0.2$ & 271 & 58 & 191 & 1,697,855 & 25.66\% & 52\\
Budget & $\alpha=0.1$ & 136 & 68 & 113 & 1,166,927 & 48.91\% & 39--40\\
Budget & $\alpha=0.05$ & 71 & 73 & 69 & 870,616 & 61.88\% & 39--40\\
\midrule
Top-3 & -- & 225 & 65 & 224 & 1,926,591 & 15.65\% & 49\\
Top-1 & -- & 75 & 75 & 75 & 920,157 & 59.71\% & 35--38\\
\bottomrule
\end{tabular}
\end{table}
\begin{table}[H]
\caption{All policies at the executed window, part 2: selection by BGE, packed in BM25 order. Columns as in \cref{tab:allpolicies}. Rank 0 is all-that-fits, the first row of that table.}
\label{tab:allpoliciesbge}
\centering\scriptsize\setlength{\tabcolsep}{4pt}
\begin{tabular}{llrrrrrl}
\toprule
Policy & Setting & Selected & Changed & Delivered & Input tokens & Saved & Correct\\
\midrule
Recall gate & rank 1 & 1,087 & 14 & 287 & 2,264,996 & 0.83\% & 50--51\\
Recall gate & rank 2 & 992 & 19 & 285 & 2,255,966 & 1.22\% & 48--49\\
Recall gate & rank 3 ($\alpha=0.1$) & 966 & 20 & 285 & 2,255,005 & 1.27\% & 49\\
Recall gate & rank 4 & 926 & 24 & 283 & 2,245,935 & 1.66\% & 48\\
Recall gate & rank 5 & 839 & 32 & 278 & 2,219,035 & 2.84\% & 45--46\\
Recall gate & rank 5--6 & 766 & 41 & 272 & 2,190,347 & 4.10\% & 47--48\\
Budget & $\alpha=0.5$ & 660 & 45 & 261 & 2,141,657 & 6.23\% & 47--48\\
Budget & $\alpha=0.5$ & 658 & 46 & 260 & 2,135,965 & 6.48\% & 47--48\\
Recall gate & rank 6--7 & 645 & 47 & 259 & 2,128,514 & 6.80\% & 48--49\\
Recall gate & rank 7--8 & 644 & 47 & 259 & 2,128,819 & 6.79\% & 48--49\\
Recall gate & rank 8--9 & 631 & 47 & 257 & 2,114,947 & 7.40\% & 50--51\\
Recall gate & rank 9--10 & 629 & 47 & 257 & 2,115,079 & 7.39\% & 49--50\\
Recall gate & rank 10--11 & 628 & 47 & 257 & 2,115,079 & 7.39\% & 49--50\\
Recall gate & rank 11--12 & 600 & 50 & 251 & 2,084,138 & 8.75\% & 47--48\\
Recall gate & rank 12--13 & 572 & 53 & 246 & 2,056,798 & 9.94\% & 48--49\\
Recall gate & rank 13--14 & 508 & 55 & 234 & 1,978,214 & 13.38\% & 49--50\\
Recall gate & rank 14--15 & 501 & 55 & 235 & 1,983,167 & 13.17\% & 49--50\\
Recall gate & rank 15--16 & 499 & 55 & 235 & 1,983,167 & 13.17\% & 49--50\\
Recall gate & rank 16--17 & 453 & 55 & 219 & 1,874,210 & 17.94\% & 50\\
Recall gate & rank 17--18 & 443 & 55 & 219 & 1,875,865 & 17.87\% & 50\\
Recall gate & rank 18--19 & 398 & 60 & 209 & 1,815,827 & 20.49\% & 53\\
Budget & $\alpha=0.3$ & 375 & 62 & 203 & 1,776,838 & 22.20\% & 54--55\\
Budget & $\alpha=0.3$ & 374 & 62 & 203 & 1,776,838 & 22.20\% & 54--55\\
Recall gate & rank 19--20 & 323 & 66 & 195 & 1,734,110 & 24.07\% & 53--55\\
Recall gate & rank 20 & 277 & 67 & 184 & 1,660,143 & 27.31\% & 48--51\\
Budget & $\alpha=0.2$ & 263 & 68 & 178 & 1,623,605 & 28.91\% & 50--52\\
Budget & $\alpha=0.2$ & 262 & 68 & 178 & 1,623,605 & 28.91\% & 50--52\\
Budget & $\alpha=0.1$ & 144 & 70 & 121 & 1,218,137 & 46.66\% & 45--46\\
Budget & $\alpha=0.05$ & 69 & 74 & 68 & 858,684 & 62.40\% & 36--38\\
\bottomrule
\end{tabular}
\end{table}

\end{document}